\documentclass{article}
\usepackage{spconf,amsmath,epsfig}
\usepackage{cite}
\usepackage{multirow}
\usepackage{multicol}

\title{SEA ICE SEGMENTATION FROM SAR DATA BY CONVOLUTIONAL TRANSFORMER NETWORKS}

\name{Nicolae-C\u{a}t\u{a}lin Ristea$^1$, Andrei Anghel$^1$, Mihai Datcu$^{1,2}$ \thanks{This work was supported by the grant of the Romanian Ministry of Education and Research, CNCS - UEFISCDI, project number PN-III-P4-ID-PCE-2020-2120, within PNCDI III.}}

\address{CEOSpaceTech, University Politehnica of Bucharest, Romania$^1$ \\ Remote Sensing Technology Institute, German Aerospace Center (DLR), Germany$^2$}

\begin{document}

\maketitle

\begin{abstract}
Sea ice is a crucial component of the Earth's climate system and is highly sensitive to changes in temperature and atmospheric conditions. Accurate and timely measurement of sea ice parameters is important for understanding and predicting the impacts of climate change. Nevertheless, the amount of satellite data acquired over ice areas is huge, making the subjective measurements ineffective. Therefore, automated algorithms must be used in order to fully exploit the continuous data feeds coming from satellites. In this paper, we present a novel approach for sea ice segmentation based on SAR satellite imagery using hybrid convolutional transformer (ConvTr) networks. We show that our approach outperforms classical convolutional networks, while being considerably more efficient than pure transformer models. ConvTr obtained a mean intersection over union (mIoU) of $63.68\%$ on the AI4Arctic data set, assuming an inference time of $120ms$ for a $400 \times 400$ km$^2$ product.
\end{abstract}


\section{Introduction}
\label{sec:intro}
Sea ice retreat, particularly in the Arctic, has been one of the most significant responses to global climate change. Therefore, sea ice cover and sea ice concentration are vital parameters for conducting climate change research and navigation in polar regions. 
To support the logistics for the transport industry, there is a high demand for local-scale high-resolution information on Arctic marine conditions. Such information is critical for operations planning, shipping routes and sustainable development of the North \cite{Carter-NCR-2018}. To this end, the research infrastructure in the ice covered areas have grown significantly in the last decades. Satellite based synthetic aperture radar (SAR) systems have been employed to monitor the vast regions of the Arctic (e.g., RADARSAT-2, RADARSAT Constellation Mission, Sentinel-1A and -1B). These systems have a high spatial resolution
and have regional coverage (e.g., up to 500 km by 500 km), making them ideal for monitoring large regions. Nevertheless, even if the satellite infrastructure assures a high amount of data, this needs to be processed and interpreted to extract key information. Considering the amount of data, manual processing is barely possible, therefore automated algorithms are needed.

Recently, deep learning models have been widely adopted for the geoscience field \cite{Ristea-ARXIV-2022, Ristea-IGARSS-2022, Ren-IGARSS-2020, Gao-GRSL-2019, Jiang-IGARSS-2022, Lattari-RS-2019}. Those models have the potential to greatly improve the efficiency and accuracy of sea ice analysis, as well as to enable the analysis of large amounts of data that would be impractical to process manually. In \cite{Ren-IGARSS-2020} the authors propose a U-Net architecture for sea ice segmentation on data acquired from Santinel-1 and manually labelled. The authors observed that even if the network was trained with a limited amount of products, the network is still able to learn the patterns and obtain high segmentation scores. Gao et al. \cite{Gao-GRSL-2019} employed a dense neural architecture to detect changes over time in the sea ice areas. They adopted a transfer learning strategy to increase the network performance. Closer to our work, in \cite{Jiang-IGARSS-2022} the authors proposed a semi-supervised algorithm based on graph convolutional networks for sea ice segmentation. They obtained superior results compared with a ResNet based architecture in the limited data scenario.

Lately, inspired by the success of self-attention layers and transformer architectures in the computer vision field \cite{dosovitskiy2020image, liu2021swin}, there have been employed transformer architectures in the geoscience field \cite{sudakow2022meltpondnet, yao2022multi, dong2021exploring, fuller2022satvit, yuan2020self} with remarkable results. For example, in \cite{sudakow2022meltpondnet} the authors use a SwinTransformer architecture, originally employed for computer vision tasks, to detect the melt ponds on Arctic sea ice. They developed a cross-channel attention into the decoder block, which boosts the model's performance. 

Distinct from all mentioned methods, we propose a hybrid convolutional transformer (ConvTr) model, which combines the benefits of convolutional networks (e.g., efficiency) and transformer blocks (e.g., global attention). The architecture has a transformer core, designed to compute attention based features in a smaller latent space. Moreover, our approach uses a large scale SAR data set, AI4Arctic \cite{ai4arctic}, enabling the network to learn general patterns from diverse scenes, instead of overfitting the training on a limited number of products.

In summary, our contribution is twofold:
\begin{itemize}
\item We propose a hybrid convolutional transformer architecture for sea ice segmentation, obtaining the best time-accuracy trade-off.
\item We train our model on a large scale data set, showing the generalisation capacity of our network.
\end{itemize}
\vspace{-0.5cm}

\begin{figure*}[!t]
\begin{center}
\centerline{\includegraphics[width=1.0\linewidth]{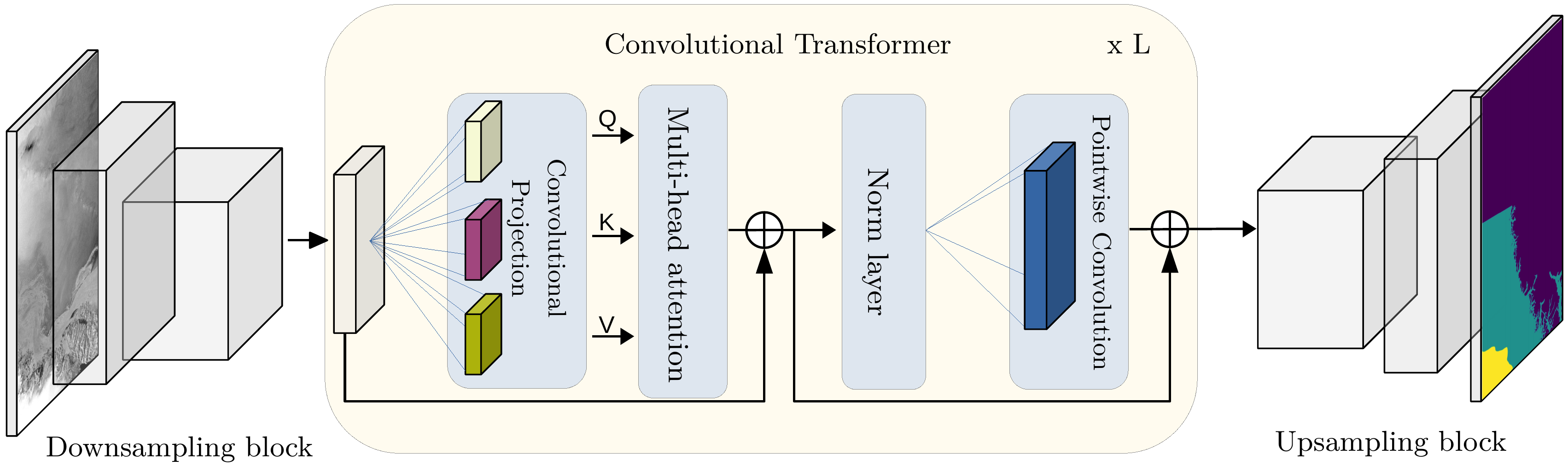}}
\caption{ConvTr segmentation architecture. The model is composed by a downsampling block comprising convolutional layers, a convolution transformer block comprising a multi-head self-attention mechanism, and an upsampling block comprising transposed convolutions.}
\label{fig_convtr}
\vspace{-1.3cm}
\end{center}
\end{figure*}

\section{Method}

\subsection{Data preprocessing}
Considering that the products from the AI4Arctic \cite{ai4arctic} data set have a high dimensional size, being impractical to train neural networks on the entire dimension, we crop fix length windows of size $P \times P$, where $P \in \mathbf{N}$ from the product. The cropping is performed such that, in each window there is more than a single label from all considered (\textit{sea}, \textit{ice}, \textit{land}). 
A training sample of size $2 \times P \times P$ contains both HH and HV polarizations. Finally, each sample is normalised before being ready to be fed into the network.

\subsection{Convolutional Transformer architecture}

In our work, we employed the ConvTr architecture composed of a convolutional downsampling block, a convolutional transformer block, and a deconvolutional upsampling block, as illustrated in Fig. \ref{fig_convtr}. We highlight that, without the convolutional downsampling block and the replacement of dense layers with convolutional layers inside the transformer block \cite{wu2021cvt}, the transformer would not be able to learn to segment images larger than $128\times 128$ pixels, due to memory overflow (measured on a Nvidia GeForce RTX 3090 GPU with 24GB of VRAM).

\noindent
\textbf{Downsampling block.}
The downsampling block starts with a convolutional layer formed of $32$ filters with a spatial support of $7 \times 7$, which are applied using a padding of $3$ pixels to preserve the spatial dimension, while enriching the number of feature maps to $32$. Next, we apply three convolutional layers composed of $32$, $64$ and $128$ filters, respectively. All convolutional filters have a spatial support of $3 \times 3$ and are applied at a stride of $2$, using a padding of $1$. Each layer is followed by batch-norm \cite{Ioffe-ICML-2015} and Rectified Linear Units (ReLU) \cite{Nair-ICML-2010}.

\noindent
\textbf{Transformer block.}
The downsampling block is followed by the convolutional transformer block, which preserves the spatial dimension between the input and output tensors. The convolutional transformer block is inspired by the block proposed in \cite{Ristea-ARXIV-2022}. More precisely, the input tensor is interpreted as a set of overlapping tokens (patches from the input tensor). The sequence of tokens is projected onto a set of weight matrices implemented as depth-wise separable convolution operations. The convolutional projection is formed of three nearly identical projection blocks, with separate parameters. The output query, keys and values (\textbf{Q}, \textbf{K}, \textbf{V}) are passed to a multi-head attention layer, with the goal of capturing the interaction among all tokens by encoding each entity in terms of the global contextual information. Next, the output passes through a batch-norm and a pointwise convolution, with the corresponding residual connections. The process is repeated $L$ times, which denotes the depth of the transformer block. The block is visually described in Fig.~\ref{fig_convtr}.

\noindent
\textbf{Upsampling block.}
The last block of our ConvTr applies upsampling operations, being designed to revert the transformation of the downsampling block. The upsampling block is formed of three transposed convolutional layers comprising $128$, $64$ and $32$ filters, respectively. All kernels have a spatial support of $3 \times 3$, being applied at a stride of $2$, using a padding of $1$. Similar to the downsampling block, we apply batch-norm and ReLU activations after each transposed convolutional layer. Finally, we employ a convolutional layer with $C \in \mathbf{N}$ filters, each filter having a kernel dimension of $7 \times 7$ and a padding of $3$, to reduce the number of channels from $32$ to $C$. In this manner, we obtained the same output dimension as the input image, where each of the $C$ channels represents the probability that a certain pixel is part of the corresponding class.

\vspace{-0.2cm}
\subsection{Loss function}

Considering the imbalanced classes from the AI4Arctic \cite{ai4arctic} data set, we optimised the model in accordance with the focal loss function \cite{lin2017focal}. In this manner, the network converged faster and was more robust at testing time to the minority class. Formally, the loss is defined below.
\vspace{-0.3cm}
\begin{equation}
\vspace{-0.3cm}
    \mathcal{L} = \alpha (1 - p_t)^{\gamma} \mathcal{L}_{CE},
\end{equation}
\noindent
where $\alpha$ control the class weights, $p_t$ is the probability of predicting the ground truth class, $\gamma$ controls the degree of down-weighting of easy-to-classify pixels and $\mathcal{L}_{CE}$ is the cross entropy loss function.

\vspace{-0.3cm}
\section{Experiments}
\vspace{-0.2cm}

\subsection{Data set}
The AI4Arctic \cite{ai4arctic} Sea Ice data set are produced for the AI4EO sea ice competition initiated by the European Space Agency. The data set contains Sentinel-1 active microwave SAR data and corresponding passive MicroWave Radiometer data from the AMSR2 satellite sensor. Each product has associated ice charts that have been produced by the Greenland ice service at the Danish Meteorological Institute and the Canadian Ice Service for the safety of navigation. The scenes are from the time period from January 8 2018 to December 21 2021. The extra wide swath GRDM products cover a region of $400 \times 400$ $km^2$, have a resolution of $90$ meters and a pixel spacing of $40$ meters. The entire data set contains 513 annotated products. We split the data into a training set (400 products) and a test set (113 products).

\begin{table}
\centering
\caption{Segmentation and inference time results on the AI4Arctic 
 \cite{ai4arctic} test set. ConvTr is compared against two baseline methods (ResNet AE, UNet \cite{ronneberger2015u}). We included for ablation the ConvTr only with convolutional blocks (AE) and only with transformer block (Transformer).}
\label{tab_results}
 \begin{tabular}{|l|cc|}
 \hline
  Method & mIoU (\%) & Inference time (ms)\\
 \hline\hline
 ResNet AE        & 53.04  & 87    \\ 
 \hline
 UNet \cite{Ren-IGARSS-2020}            & 56.43  & 92    \\
 \hline
 \hline
 AutoEncoder                 & 49.75  & 65    \\ 
 \hline
 Transformer        & 63.81  & 473    \\ 
 \hline
 ConvTr (ours)            & 63.68  & 120    \\ 
 \hline

\end{tabular}
\end{table} 

\vspace{-0.2cm}
\subsection{Hyper-parameters tuning}
ConvTr is optimised with Adam using the focal loss function \cite{lin2017focal}. We start with an initial learning rate of $10^{-4}$ and use a decay factor of $0.5$ after every 10 epochs. We train each model for 50 epochs on mini-batches of 16 samples. We set the number of blocks to $L\!=\!5$ and each block has $5$ attention heads. Regarding the training patch size, we found the optimal value to be $P = 512$.

\vspace{-0.2cm}
\subsection{Evaluation metrics}
Since we perform semantic segmentation between three classes, we found the most insightful metric to be mean intersection over union (mIoU). The metric captures the overall performance of the models, regardless of the unbalanced distribution between classes. In addition, we reported the inference time for a full resolution scene, which has a spatial dimension about $1100 \times 1100$.

\subsection{Results}
In Table \ref{tab_results} we report the results for ConvTr against two baseline methods, ResNet based auto-encoder and UNet \cite{ronneberger2015u}, used in \cite{Ren-IGARSS-2020}.
We observe that ConvTr surpass with more than $7\%$ both baseline methods, while marginally raising the inference time. Regarding the importance of different parts of the network, when we only use the downsampling and upsampling blocks, we have the best inference speed, but the performance is drastically affected. If we employ only the transformer block, we note that the speed is highly impacted, while the accuracy is with only $0.13\%$ higher. Therefore, combining both architectures, we exploit the benefits of those, attaining the best performance-speed trade-off.

\begin{figure}[]
\begin{center}
\centerline{\includegraphics[width=1.0\linewidth]{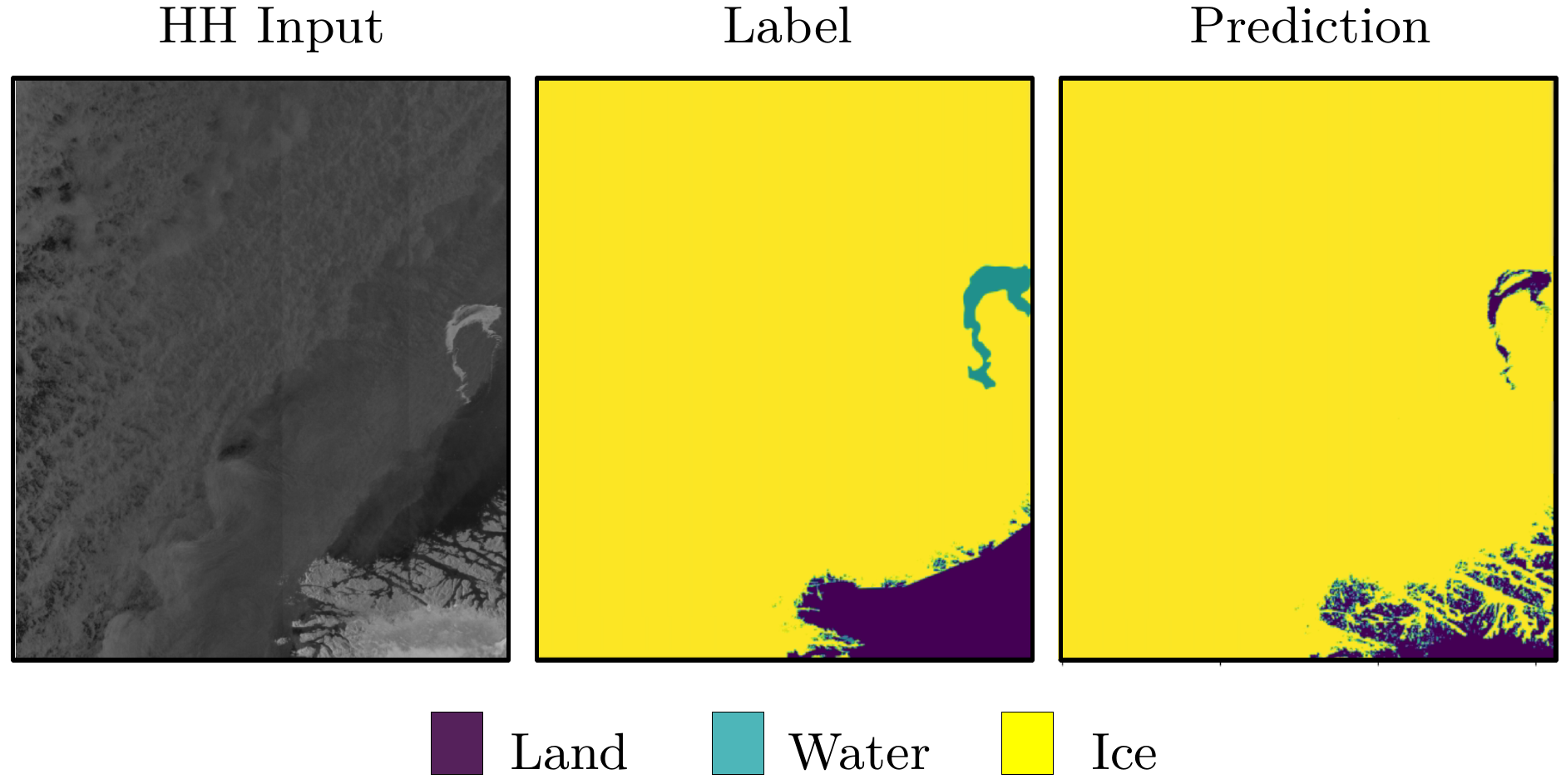}}
\caption{Results obtained with ConvTr model for product \textit{20180607T184326}. Along with the prediction, we also included the HH input and the label.}
\label{fig_res}
\vspace{-1.0cm}
\end{center}
\end{figure}

In addition to the objective metrics, we included in Fig.~\ref{fig_res} the result of our best ConvTr on \textit{20180607T184326} product from the test set. We observe that the ice and land classes are well segmented, while the water class is miss classified. A potential reason could be the imbalanced training set.

\section{Conclusion}

In this paper, we propose a hybrid convolutional transformer architecture for sea ice segmentation, based on SAR data. We trained our model on a large scale data set, testing the generalisation capacity on over 100 products. Moreover, we showed that our hybrid architecture attains the best performance-speed trade-off, being feasible to be deployed for automated segmentation.

\vspace{-0.25cm}

\bibliographystyle{IEEEbib}
\bibliography{main}

\end{document}